\documentclass[runningheads]{llncs}
\usepackage{graphicx}
\graphicspath{{./figs/}{./figs}}
\usepackage{comment}
\usepackage{amsmath,amssymb} 
\usepackage{color}
\usepackage{multirow}
\usepackage{array}

\begin{document}
\pagestyle{headings}
\mainmatter
\def\ECCVSubNumber{3152}  

\title{Unsupervised 3D Human Pose Representation with Viewpoint and Pose Disentanglement} 

\titlerunning{Unsup. 3D Human Pose with Viewpoint and Pose Disentanglement}
%
\author{Qiang Nie\inst{1,2}~\orcidID{0000-0002-2778-4058} \and
Ziwei Liu\inst{1}~\orcidID{0000-0002-4220-5958} \and \\
Yunhui Liu\inst{1,2}~\orcidID{0000-0002-3625-6679}}
\authorrunning{Nie et al.}
%
\institute{
The Chinese University of Hong Kong, Shatin N.T., Hong Kong \and
T Stone Robotics Institute of CUHK \\
\email{\{qnie,yhliu\}@mae.cuhk.edu.hk, zwliu@ie.cuhk.edu.hk}
}
\maketitle

\begin{abstract}
Learning a good 3D human pose representation is important for human pose related tasks, \textit{e.g.} human 3D pose estimation and action recognition. Within all these problems, \textit{preserving the intrinsic pose information} and \textit{adapting to view variations} are two critical issues. In this work, we propose a novel Siamese denoising autoencoder to learn a 3D pose representation by disentangling the pose-dependent and view-dependent feature from the human skeleton data, in a fully unsupervised manner. These two disentangled features are utilized together as the representation of the 3D pose. To consider both the kinematic and geometric dependencies, a sequential bidirectional recursive network (SeBiReNet) is further proposed to model the human skeleton data. Extensive experiments demonstrate that the learned representation 1) preserves the intrinsic information of human pose, 2) shows good transferability across datasets and tasks. Notably, our approach achieves state-of-the-art performance on two inherently different tasks: pose denoising and unsupervised action recognition. Code and models are available at: \url{https://github.com/NIEQiang001/unsupervised-human-pose.git}.

\end{abstract}

\section{Introduction}
\label{intro}

Human action recognition and human behavior analysis have extensive applications on human-robot interaction (HRI) systems, such as health caring, entertainment, education, security and many other intelligent surveillance scenarios, which also makes the 3D human pose estimation a hot research topic for many decades. Learning a good human 3D pose representation has great significance both to the research of human action recognition and the human pose estimation.

While understanding the human pose is a challenging task, which requires the computer to learn the dependencies between joints of the human skeleton robustly in different viewpoints. These dependencies include kinematic relationships between joints and geometric features of the human body. The kinematic relationship describes the motion transmission process between joints and the role of each joint in an action. The geometric feature refers to those specific appearance characteristics of the human body, such as fixed bone lengths and the symmetry between left and right limbs. Many existing works have utilized the geometric features of the human body~\cite{ramakrishna2012reconstructing,liu2016fashion,sun2017compositional,zhou2017weaklysupervised,rong2019delving}, but few works are capable to model the kinematic relationships between human body joints. Kinematics is a physical process and hard to be modeled by regular CNN, RNN or MLP neural networks. Hence, we proposed a sequential bidirectional recursive network (SeBiReNet) to model the dependencies of the human skeleton.

Besides the dependencies between joints, the human 3D pose presents infinite modalities when recorded or observed from different viewpoints, which makes the processing of the human 3D pose quite intractable for the intelligent system. Increasing the size of training dataset with different views may be effective. However, it's impossible to record the data from all possible viewpoints. To tackle the view variation, some previous works applied preprocessing treatment to eliminate the view variation~\cite{liu2017enhanced,demisse2018pose}. These methods are always dataset dependent because of the specifically designed preprocessing method. Many other methods~\cite{yang2012eigenjoints,nie2019view,yang2012eigenjoints,huang2017deep,vemulapalli2014human,xia2012view} extracted hand-crafted view-invariant features as pose descriptors based on the prior knowledge of human beings. Although these hand-crafted features are view-invariant, there is information loss in extracting these features as only a few explanatory factors are considered. There are some methods~\cite{demisse2018pose,li2018unsupervised,zheng2018unsupervised} trying to learn discriminative pose representations using the deep learning method. However, the transferability of the representations learned by existing approaches in different datasets and different tasks is limited.

Human pose result from the rich interaction of many factors, such as the subject, the action, and the viewpoint. Learning view-invariant features means to extract features that are insensitive to the direction of view variation, which also means some features that are sensitive to the variations but informative are discarded. As Bengio et al.~\cite{bengio2013representation} mentioned, a better way to overcome these challenges is to leverage the data itself, ..., to disentangle as many factors as possible and discarding as little information about the data as in practice.

Motivated by aforementioned issues, we propose an unsupervised method for learning a latent representation of the human 3D pose by disentangling the pose-dependent and view-dependent features from human skeleton data. We introduce a novel SeBiReNet to model the human skeleton data. A Siamese denoising autoencoder based on the SeBiReNet is designed to learn the latent human pose representation. Ability of denoising corrupted skeletons from an unseen dataset proves the learned representation preserves the intrinsic information of human pose, including both the kinematic and geometric dependencies. Disentangling the pose-dependent (view-invariant) and view-dependent (view-variant) feature from skeleton data other than extracting the view-invariant feature enables us to transfer the viewpoint of human pose in the latent space, which is used as a strengthened regularization in our training process. 

We summarize our contributions as follows:
\begin{itemize}	
	\item[$\bullet$] We propose a novel SeBiReNet to model the kinematic dependencies between body joints in the human skeleton data.
	
	\item[$\bullet$] Based on SeBiReNet, a Siamese denosing autoencoder is proposed for learning 3D human pose representation with feature disentanglement.	
	The unsupervisedly learned pose representation 1) preserves the intrinsic information of human pose, 2) shows good transferability across datasets and tasks. 
	
	\item[$\bullet$] Extensive experiments demonstrate that state-of-the-art performance can be achieved when applying the learned representation on two inherently different tasks: pose denoising and unsupervised action recognition.
\end{itemize}

\section{Related Works}
\label{related_works}

\subsection{Modeling Human 3D Poses} 
%
To understand the human 3D pose, the most important is to figure out the dependencies between body joints, which should include both the kinematic and the geometric dependency. Compared to kinematic dependency between joints, geometric characteristic is much easier to model. Ramakrishna et al.~\cite{ramakrishna2012reconstructing} used normalized limb lengths as anthropometric regularity to reconstruct 3D human pose from the 2D image landmarks. Sun et al.~\cite{sun2017compositional} proposed to use the summation of bone lengths as a supervision loss. The summation of bone length considers all bones between every paired two joints. In essence, the summation of bone lengths is a pairwise geodesic distance. Their work proved that the accuracy of human pose regression can be improved based on the summations of bone length. As ratios between bone lengths remain relatively fixed in a human skeleton, Zhou et al.~\cite{zhou2017weaklysupervised} utilized the length ratios of bones as a weak supervision for reconstructing 3D pose from wild images without 3D annotations. Though human skeleton is similar to the tree structure, few works have applied the recursive network for the human pose modeling. Wei et al.~\cite{wei2017human} introduced a vanilla tree network for skeleton-based action recognition. However, only the output from the single tree root node is utilized, which is inherently different from the structure of our SeBiReNet proposed to model the human 3D pose.

\subsection{Learning Pose Representations} 
Demisse et al.~\cite{demisse2018pose} proposed a denoising autoencoder for unsupervised skeleton-based action recognition by using MLP layers. But to evaluate the extracted features in cross-view action recognition, a preprocessing treatment is applied to estimate the view variation. Li et al.~\cite{li2018unsupervised} proposed a method to learn pose representation from sequential RGB data by adding a view discriminator to decide which view the learned feature comes from. Using view classifier indicates that their views are depend on the training data and view labels were given. While in our method, the poses are randomly rotated and no view label is given. Zheng et al.~\cite{zheng2018unsupervised} presented an adversarial training strategy to learn representations of skeleton sequences for action recognition. Compared to these methods, the proposed method is able to learn a view-invariant pose-dependent feature from single pose without any additional label or auxiliary network. Requiring no temporal information makes our representation can be applied to both time-related or time-independent tasks, as verified in our experiments. It's interesting to find that Aberman et al.~\cite{aberman2019learning} applied a similar feature decomposition and re-composition process in their work of retargeting video-captured motion between different human performers. Our method differs with theirs in two aspects: 1) we embed a denoising process in the learning, which helps the network capture the intrinsic feature of skeleton pose; and 2) our disentangled features have more interaction by sharing some weights in the decomposition and multiplying with each other in the re-composition process.

\section{Our Approach}

\subsection{Problem Formulation}
\label{PF}
Given a human 3D pose $x$, a latent representation $h$ can be learned by assigning a function $f$ with parameters $\theta$, i.e., $h=f(x;\theta)$. In order to make sure the learned representation contains useful information of original data, $h$ is required to be able to recover the original pose through another function $g$ with parameters $\zeta$. The reconstructed pose can be formulated as $\hat{x}=g(h;\zeta)$, which is the basic idea of autoencoder in representation learning. Vincent et al.~\cite{vincent2008extracting} has proven that using the denoising autoencoder to reconstruct the clean data from its corrupted version is helpful in avoiding trivial solutions and improving the performance of learned latent representations. Therefore, basically, learning the human 3D pose representation can be formulated as the following equation.

\begin{equation} \label{e1}
\mathop{\arg\min}_{\theta, \zeta} \mathbb{E}_{q(\tilde{x}|x)}L(x, g(h; \zeta))
\end{equation}

\noindent where $h=f_{\theta}(\tilde{x})$ is the learned latent representation of the human 3D pose and $\tilde{x}$ is the corrupted pose corresponding to the clean pose $x$. However, the representation learned in eq.~\ref{e1} contains both the pose-dependent and the view-dependent information. As analyzed in Sec.~\ref{intro}, we hope to learn a view-invariant representation as well as avoid discarding the view-dependent feature of human pose for the sake of information preservation. Thus different from traditional methods, we attempt to disentangle the view-invariant feature $h_{vi}$ from view-dependent feature $h_v$ and using the combination of $\left[ h_{vi}, h_v \right] $ as a latent representation of the human 3D pose. Under this consideration, the representation learning is reformulated as eq.~\ref{e2}.

\begin{equation} \label{e2}
\mathop{\arg\min}_{\theta_{vi},\theta_v,\zeta} \mathbb{E}_{q(\tilde{x}|x)}L(x, g(h_v \otimes h_{vi}; \zeta))
\end{equation}

\noindent where $h_{vi}=f(\tilde{x}; \theta_{vi})$ and $h_v=f(\tilde{x}; \theta_v)$. $\otimes$ is an operation to couple $h_{vi}$ and $h_v$ together, which can be matrix multiplication or concatenation. From a generative perspective, the learning process in eq.~\ref{e2} can also be written as 

\begin{equation} \label{e3}
\mathop{\arg\max}_{\theta_{vi},\theta_v, \zeta} \mathbb{E}_{q(\tilde{x}|x)}\log \left[ p(x|h_{vi}, h_v;\zeta)p(h_{vi},h_v|\tilde{x};\theta_{vi}, \theta_v)p(\theta_{vi},\theta_v)\right] 
\end{equation}

\noindent where $q(\tilde{x}|x)$ denotes the pose corruption process. If we assume the prior distribution $p(\theta_{vi},\theta_v)$ can be factorized as $p(\theta_{vi},\theta_v)=p(\theta_{vi})p(\theta_v)$, i.e., they are independent. Then we have
\begin{equation} \label{e4}
\log p(h_{vi},h_v|\tilde{x};\theta_{vi}, \theta_v)p(\theta_{vi},\theta_v)
=\log p(h_{vi}|\tilde{x};\theta_{vi})p(\theta_{vi}) + \log p(h_v|\tilde{x};\theta_v)p(\theta_v)
\end{equation}

According to eq.~\ref{e4}, learning of the view-dependent feature and pose-dependent feature don't have much influence on each other. To strengthen the interaction between these two features and disentangle them smoothly, we propose to have $p(h_{vi}|\tilde{x};\theta_{vi})=f(\phi(\tilde{x};\eta);\theta_{vi} \backslash \eta)$ and $p(h_v|\tilde{x};\theta_v)=f(\phi(\tilde{x};\eta);\theta_v \backslash \eta)$, where $\eta$ is the shared parameters in the parameter space. In this manner, $h_v$ and $h_{vi}$ are disentangled and affect each other through the common latent feature $\phi(\tilde{x})$. Although we are trying to disentangle the view-dependent and pose-dependent feature from original pose, this is not necessarily induced so far, as the learned latent representation $[h_{vi}, h_v]$ is still not well constrained. To introduce the concept of viewpoint into the learning process, an additional transformation is added to the corrupted pose $\tilde{x}$ by randomly rotating it in the 3D space. At this circumstance, eq.~\ref{e3} becomes 
\begin{equation} \label{e5} 
\mathop{\arg\max}_{\theta_{vi},\theta_v, \zeta} \mathbb{E}_{q(\tilde{x}|x)q_r(\tilde{x}_r|\tilde{x})}\log\left[ p(x|h_{vi}, h_v;\zeta)p(h_{vi},h_v|\tilde{x}_r;\theta_{vi}, \theta_v)p(\theta_{vi},\theta_v)\right] 
\end{equation}

\noindent where $\tilde{x}_r$ is the randomly rotated corrupted pose corresponding to $\tilde{x}$, $q_r(\tilde{x}_r|\tilde{x})$ denotes the random rotation process. A marginal distribution consistency of $p(h_{vi})$ should be satisfied from corrupted pose $\tilde{x}$ and randomly rotated pose $\tilde{x}_r$. Thus, besides the pose reconstruction loss, we regularize the pose-dependent feature by minimizing the Kullback\textendash Leibler divergence between pose-dependent features of poses under different observation angles as shown in eq.~\ref{e6}.

\begin{equation} \label{e6}
\mathop{\arg\min}_{\theta_{vi},\theta_v} D_{KL}(p(h_{vi}|\tilde{x};\theta_{vi})||p(h_{vi}|\tilde{x}_r;\theta_{vi}))
\end{equation}

Putting all together, our human 3D pose representation learning process is modelled as
\begin{multline} \label{e7}
\mathop{\arg\min}_{\theta_{vi},\theta_v, \zeta} \mathbb{E}_{q(\tilde{x}|x)}\left[ L(x, g(h_v \otimes h_{vi}, \tilde{x}; \zeta)) + q_r(\tilde{x}_r|\tilde{x})L(x, g(h_v \otimes h_{vi}, \tilde{x}_r; \zeta)) \right] + \\ D_{KL}(p(h_{vi}|\tilde{x};\theta_{vi})||p(h_{vi}|\tilde{x}_r;\theta_{vi}))
\end{multline}

\subsection{Sequential Bidirectional Recursive Network}
\label{SeBiReNet}

\begin{figure}[t!]
	\begin{center}
		\includegraphics[width=0.55\linewidth]{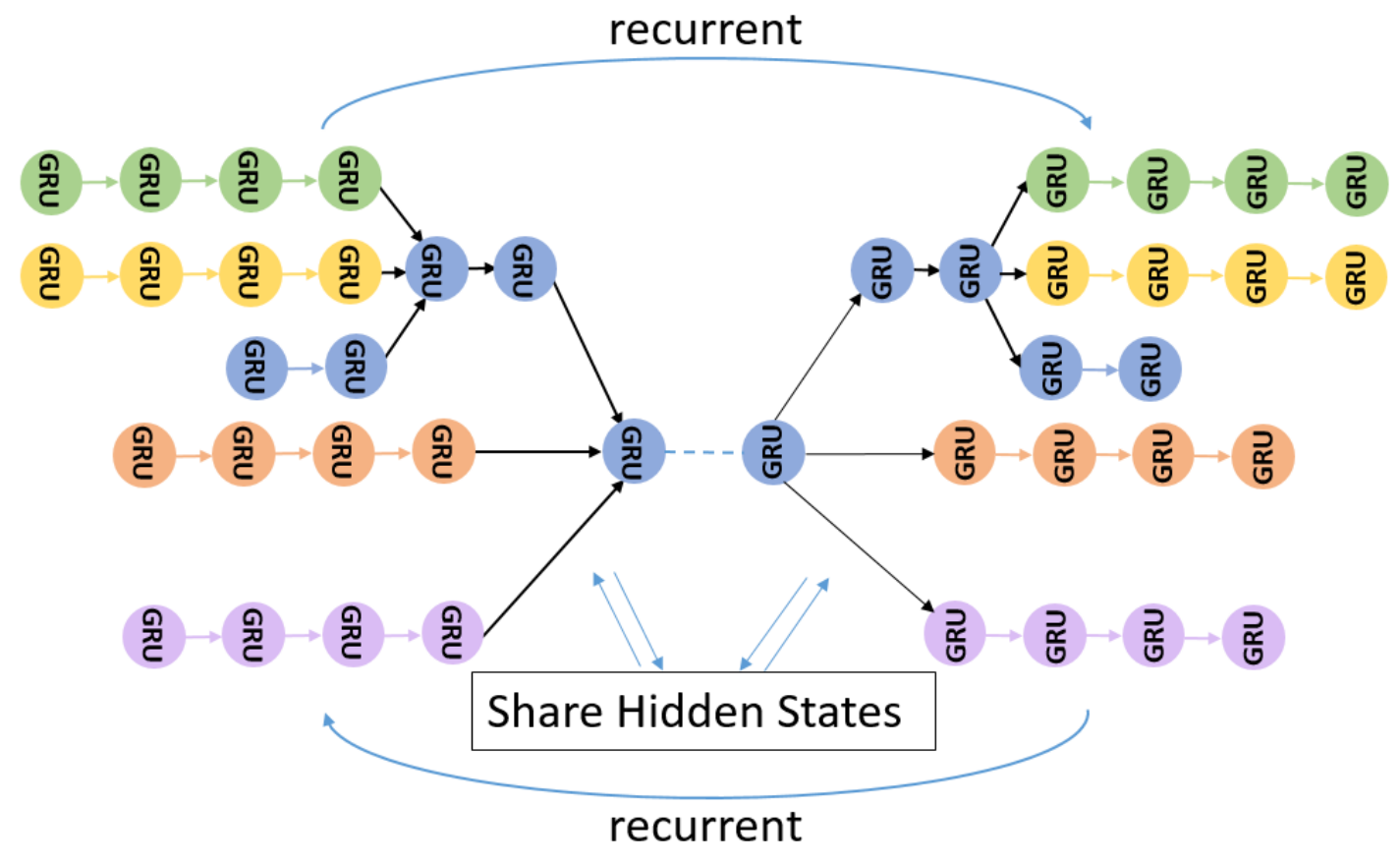}
	\end{center}
	\caption{The proposed sequential bidirectional recursive neural network (SeBiReNet). Each node corresponds to a real joint of the human body and different colors represent different body parts}
	\label{fig:SeBi}
\end{figure}

In order to capture the kinematic dependencies of human skeleton structure, a sequential bidirectional recursive neural network (SeBiReNet) is proposed. The bidirectional recursive neural network has two tree structures as shown in Fig.~\ref{fig:SeBi}, which models the human skeleton structure intuitively. The recursive neural network is widely used for text or language analysis~\cite{irsoy2014deep,socher2010learning} due to its ability in summarising the semantic meanings. However, the conventional recursive neural network has only one direction, which means the information can only flow from leaf nodes to the root node. On the contrary, the motion of the human body is transmitted from parent joint to child joints. Usually, to determine the position of a joint, both the position of parent joint and the positions of child joints have to be considered. In this regards, the proposed SeBiReNet models the dependency $p(J_{parent}|J_{child})$ and $p(J_{child}|J_{parent})$ between parent joint and child joint respectively through a recursive subnet (left part in Fig.~\ref{fig:SeBi}) and a diffuse subnet (right part in Fig.~\ref{fig:SeBi}). The two subnets have independent kernel weights but share the hidden states $h\in \mathbb{R}^{J\times m}$, where $J$ is the joint number and $m$ is the feature size. The shared hidden states store the intermediate inference results when information flows in the network, and the intermediate results will be continually refined when the network recurrently runs. This network is named SeBiReNet because information flows sequentially and reversely in the two subnets. The proposed architecture not only models the forward and inverse kinematic process but also imitates the repeated thinking process of human.

The node number of SeBiReNet can be adjusted according to the joint number of a human skeleton model. As most skeleton models contain 17 joints, the basic version of our proposed model is designed to have 34 nodes. In SeBiReNet, each node is a GRU cell. Other node types, such as LSTM, can also be used. The forgetting mechanism GRU cell enables the network to tackle noisy input. The inference process of SeBiReNet can be formulated as equation~\ref{e8}.
\begin{equation} \label{e8}
\begin{split}
&{h^r_i} = \varphi({W^r_{xi}}x^r_i + W^r_{hi}h{_i} + b^r_{i})\\
&O^r_i = \mathcal{O}(W^r_{o}h^r_i + b^r_o)\\
&{h^d_i} = \varphi({W^d_{xi}}x^d_i + W^d_{hi}h{_i} + b^d_{i})\\
&O^d_i = \mathcal{O}(W^d_{o}h^d_i + b^d_o)\\
\end{split}
\end{equation} 
\noindent where $x^r_i = (p_i, h_{children})$ and $x^d_i = (p_i, h_{parent})$ are the input of the node $i$, which contains the 3D position $p_i$ of corresponding joint $i$ and the hidden states output from all its child nodes $h_{children}$ or parent node $h_{parent}$. $h_i \in \mathbb{R}^m$ denotes the shared hidden state of the node $i$. The superscript $r$ represents the recursive subnet and $W^r_{xi}, W^r_{hi}, b^r_{i}, W^r_{o}, b^r_o$ are kernel weights and biases of it. The superscript $d$ denotes the diffuse subnet and $W^d_{xi}, W^d_{hi}, b^d_{i}, W^d_{o}, b^d_o$ are kernel weights and biases belong to it. $\varphi$ denotes the nonlinear function of the GRU cell. $\mathcal{O}$ is the activation function and $tanh$ is used in our experiments. After each inference in the recursive subnet or diffuse subnet, the shared hidden states and network output will be updated by the hidden states and output of corresponding subnet. Outputs of all nodes are concatenated together as the final output of the SeBiReNet.

\textbf{Complexity of the SeBiReNet} Assuming the hidden units of each GRU node is $n_h$ and the dimension of input feature is $n_x$. The number of parameters in a node with $N$ child nodes (in recursive subnetwork) or parent node (in diffuse subnetwork) is $l_N=\left[ 3n_x +(3N+4)n_h+1\right] n_h$. In a SeBiReNet with 17 joints, there are 6 leaf nodes ($l_0$), 26 nodes with one child node or parent node ($l_1$), 2 nodes have 3 child nodes ($l_3$).

\subsection{Learning Framework Based on SeBiReNet}
\label{learning_architecture}

According to the analysis in Sec.~\ref{PF}, we designed a denoising autoencoder (DAE) to learn the representation of the human 3D pose based on SeBiReNet. Different from general practice~\cite{vincent2010stacked} that adds Gaussian noise to the clean input and achieves a gently polluted version, we directly destroy the skeleton to an unreasonable version where some randomly selected joints are moved to illegal positions. The network is expected to distinguish valid human pose from invalid human pose and recover the correct position of those invalid joints.

\begin{figure*}[t!]
	\begin{center}		
		\includegraphics[width=\linewidth]{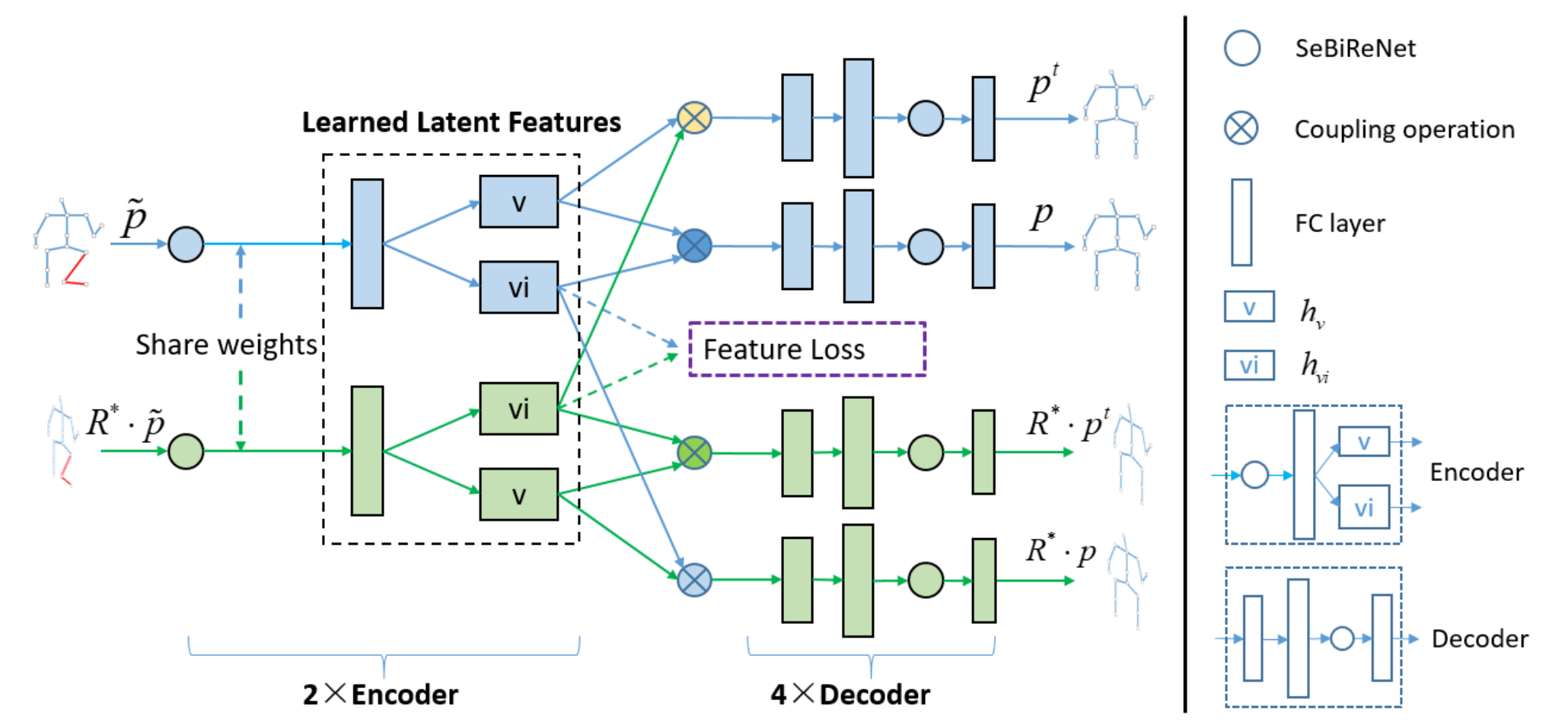}
	\end{center}
	\caption{The proposed architecture for the human 3D pose representation learning, which takes randomly corrupted 3D skeletons as inputs and reconstructs their correct version. Blue stream processes the non-rotated poses and green stream processes the randomly rotated poses}
	\label{fig:wholearchitecture}
\end{figure*}

Though the kinematic dependency has intrinsically modeled by the SeBiReNet, the geometric characteristics haven't been well considered. To this end, we added a bone length loss $L_B$ and a symmetry loss $L_S$ to the pose reconstruction loss, as shown in eq.~\ref{e9}. 
\begin{equation}\label{e9}  
\begin{split}
L(p, g(h_v \otimes h_{vi}, \tilde{p};\zeta))= \sum_{s}\left( L_P + L_B + L_S \right)
\end{split}
\end{equation}
\noindent where the first part $L_P=\sum_{i=1}^{J}\|p^s_i - \hat{p}^s_i)\|_2$ is the reconstruction error of joint position, $p^s_i$ denotes the 3D position of joint $i$ of sample $s$, $\hat{p}^s_i$ is the corresponding recovered position. The second term $L_B=\sum_{ij}\|b^s_{ij} - \hat{b}^s_{ij}\|_2$ calculates the bone length loss, which requires the recovered bone length $\hat{b}^s_{ij}$ between joint $i$ and $j$ to be equal to the ground truth length $b^s_{ij}$. The third term $L_S=\sum_{mn, kl}\|\hat{b}^s_{mn} - \hat{b}^s_{kl}\|_2$ constrains the recovered bone lengths of the left limb must be equal to the corresponding bone lengths of recovered right limb. 

The view-dependent feature and pose-dependent feature are disentangled after the SeBiReNet in the encoder. Sharing some weights before disentanglement can strengthen the interaction between $h_v$ and $h_{vi}$ as explained in Sec.~\ref{PF} and make the network more compact. It's a reasonable requirement that view-dependent feature should not change the metrics of pose-dependent feature space. As the coupling operation $\otimes$ we adopt is matrix multiplication, the requirement is satisfied only when the view-dependent feature plays a role of unit unitary transformation. For our real domain problem, we regularize the view-variant feature $h_v \in \mathbb{R}^{z \times z}$ in the $SO(z)$ space as shown in eq.~\ref{e10}, where $z \times z$ is the dimension of $h_v$. $\lambda$ is a weight factor and $I$ is an identity matrix. The orthogonal regularization is also capable of preventing the pose-related information from leaking into view-dependent feature.
\begin{equation} \label{e10}
L_{O} = \lambda\|I - h_v^Th_v\|_2 
\end{equation}
To regularize the learned pose-dependent feature being view-invariant, random rotation is added to those corrupted human poses and keeping consistency between distribution $p(h_{vi}|\tilde{p})$ and $p(h_{vi}|R^* \cdot \tilde{p})$ by using a feature loss, as shown in Fig.~\ref{fig:wholearchitecture}. In Fig.~\ref{fig:wholearchitecture}, there are two pipelines to process the corrupted pose $\tilde{p}$ and the randomly rotated pose $R^*\cdot \tilde{p}$ separately. $R^*$ is a randomly generated rotation matrix. The SeBiReNet is utilized both in the encoder and decoder. Weights are shared between all the encoders and decoders to make sure that poses under different views are encoded and decoded in the same manner. The feature loss $L_{f}$ of learned pose-dependent features from different views is defined as the Frobenius norm $L_{f}=\|h^1_{vi} - h^2_{vi}\|_F$. We believe that, if features are well disentangled from human pose, poses can be transfered between different views by exchanging their pose-dependent features and view-dependent features. This belief is utilized as a reinforced regularization for learning the pose representation in our method, as shown in Fig.~\ref{fig:wholearchitecture} where $p^t$ and $R^*p^t$ are view-transferred poses. Therefore, writing all together, our optimization target of learning a human 3D pose representation is formulated as eq.~\ref{e11}, where $L(p), L(R^*p), L(p^t), L(R^*p^t)$ are the pose reconstruction loss defined in eq.~\ref{e9}, $\omega_1, \omega_2, \omega_3$ are weights to adjust the influence of each loss, $R(w)$ is the L2 weight regularization term to avoid overfitting.
\begin{equation}\label{e11}
\mathop{\arg\min}_{\theta_{vi}, \theta_v, \zeta} \left\lbrace L(p) + L(R^*p)+ \omega_1 L(p^t) + \omega_2 L(R^*p^t) + \omega_3 L_{f} + L_{O} + R(w) \right\rbrace  
\end{equation}

\section{Experiments}
\subsection{Experimental Setup}
\begin{figure*}[t!]
	\centering
	\includegraphics[width=\linewidth]{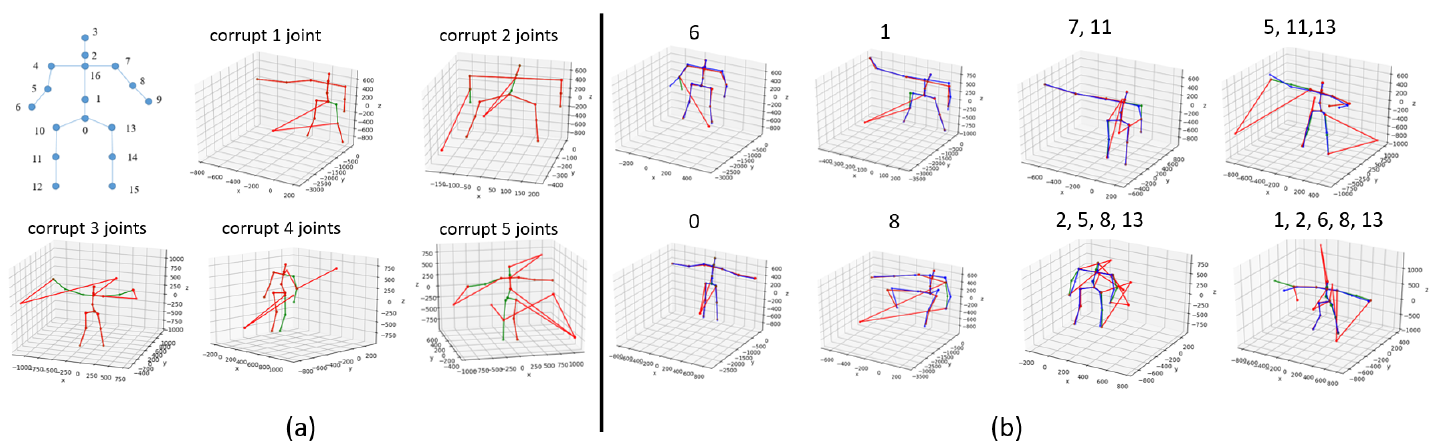}	
	\caption{(a) Illustration of the skeleton model and some generated corrupted skeleton samples, (b) Pose recovery results from randomly corrupted skeletons, the above number notes the id of corrupted joint(s). The green line, red line and blue line draw the ground truth skeleton, the corrupted skeleton and recovered skeleton, respectively. Better to view in color mode with scaling}
	\label{fig:APEsm}       
\end{figure*}

\noindent
\textbf{Implementation Details.}
The hidden unit number of GRU cell in SeBiReNet is 32. Except the output layer, nonlinear activation function $tanh$ is utilized after each MLP layer. Gradient descent optimizer with an initial learning rate of 5e-5 is used in training the DAE. Weights of different losses defined in eq.~\ref{e11} are $\omega_1=0.01, \omega_2=0.01, \omega_3=0.1$. $\lambda$ in eq.~\ref{e10} is set to 0.1. The batch size is 64. 

\noindent
\textbf{Training Set.}
The Cambridge-Imperial APE (Action-Pose-Estimation) dataset is used to train the proposed Siamese DAE. The Cambridge-Imperial APE dataset, which contains 245 sequences from 7 subjects performing 7 different categories of actions, is collected for 3D human pose estimation. Corrupted skeletons are generated by randomly selecting 1$\sim$5 joints from each skeleton and moving the them to unreasonable positions with a relatively large displacement. As shown in Fig.~\ref{fig:APEsm} (a), these corrupted skeletons violate bio-constraints, such as bone length and allowed motion angle limits. Totally, 52500 corrupted poses are generated for training and 14000 skeletons are generated for testing. The Mean Per Joint Position Error (MPJPE) is adopted as a performance measurement of reconstructed skeletons and the trained model.

\noindent
\textbf{Test Sets.}
To verify the effectiveness of learned representations, we evaluate them on two different tasks: pose denoising and unsupervised cross-view action recognition. Two benchmark action datasets are used: Northwestern-UCLA (N-UCLA) dataset~\cite{wang2014cross} and NTU RGB+D dataset~\cite{Shahroudy_2016_NTURGBD}. Both of the two datasets contain skeletons captured from different views and performed by different subjects. NTU RGB+D dataset is one of the largest skeleton datasets and N-UCLA is one of the most commonly used datasets. Pretrained encoder is applied on them to extract pose representations without any additional training, i.e., these two datasets are not used in the training phase of DAE. A 1-layer LSTM with 128 hidden units is used as the classifier in action recognition task.  

\begin{table}[t!]
	\centering
	\caption{Comparison of the performance on pose denoising among different network structures. The proposed structure achieves the best results}
	\label{tab:1}       
	\begin{tabular}{lcl}
		\hline\noalign{\smallskip}
		Network Structure & MPJPE (mm)  \\
		\noalign{\smallskip}\hline\noalign{\smallskip}
		conventional tree (only has the recursive subnet)~\cite{wei2017human} & 65.76 \\
		the diffuse subnet & 64.94 \\
		concatenated structure & 75.17 \\
		\textbf{SeBiReNet} & \textbf{42.03} \\
		\textbf{recurrent SeBiReNet} &   \textbf{41.58} \\
		\noalign{\smallskip}\hline
	\end{tabular}
\end{table}

\subsection{Evaluation on Pose Denoising}
Our model is trained and validated on the Cambridge-Imperial APE dataset. Fig.~\ref{fig:APEsm} (b) shows several recovered skeletons. Although we destroy the skeleton randomly and extremely, our network is still able to recover the correct positions of those invalid joints. To further show the effectiveness of our network design, we compared the performance of the proposed SeBiReNet with some baseline structures: conventional tree structure (only has the recursive part), the diffuse subnet, the concatenated structure, and the recurrent SeBiReNet. Different from the SeBiReNet which shares hidden states between the recursive subnet and the diffuse subnet, the concatenated structure takes the concatenation of the outputs from the recursive subnet and the diffuse subnet as its output. The recurrent SeBiReNet means the SeBiReNet runs in a recurrent mode as shown in Fig~\ref{fig:SeBi}. In this experiment, we only implement it one more times.

For a fair comparison of the capability of different structures in encoding the human 3D pose, results in Table~\ref{tab:1} is achieved by replacing the the decoder in Fig.~\ref{fig:wholearchitecture} with a three-layer MLP. As shown in Table~\ref{tab:1}, even with a simple decoder, using the proposed SeBiReNet as encoder achieves an MPJPE of 42.03 mm, which is a 35\% improvement compared to the first three structures in recovering corrupted skeletons. Recurrently running the SeBiReNet doesn't bring too much promotion. As skeleton data is relatively simple and low dimension, implementing the SeBiReNet only once is enough to obtain a good result. Compared to structures that only has SeBiReNet in encoder, the proposed structure in Fig.~\ref{fig:wholearchitecture} which embeds the SeBiReNet both in encoder and decoder attains the best performance 33.39mm. The noteworthy result indicates that the SeBiReNet is superior to MLP layers in processing skeleton data.

\begin{figure}[t!]
	\centering
	\includegraphics[width=0.9\linewidth]{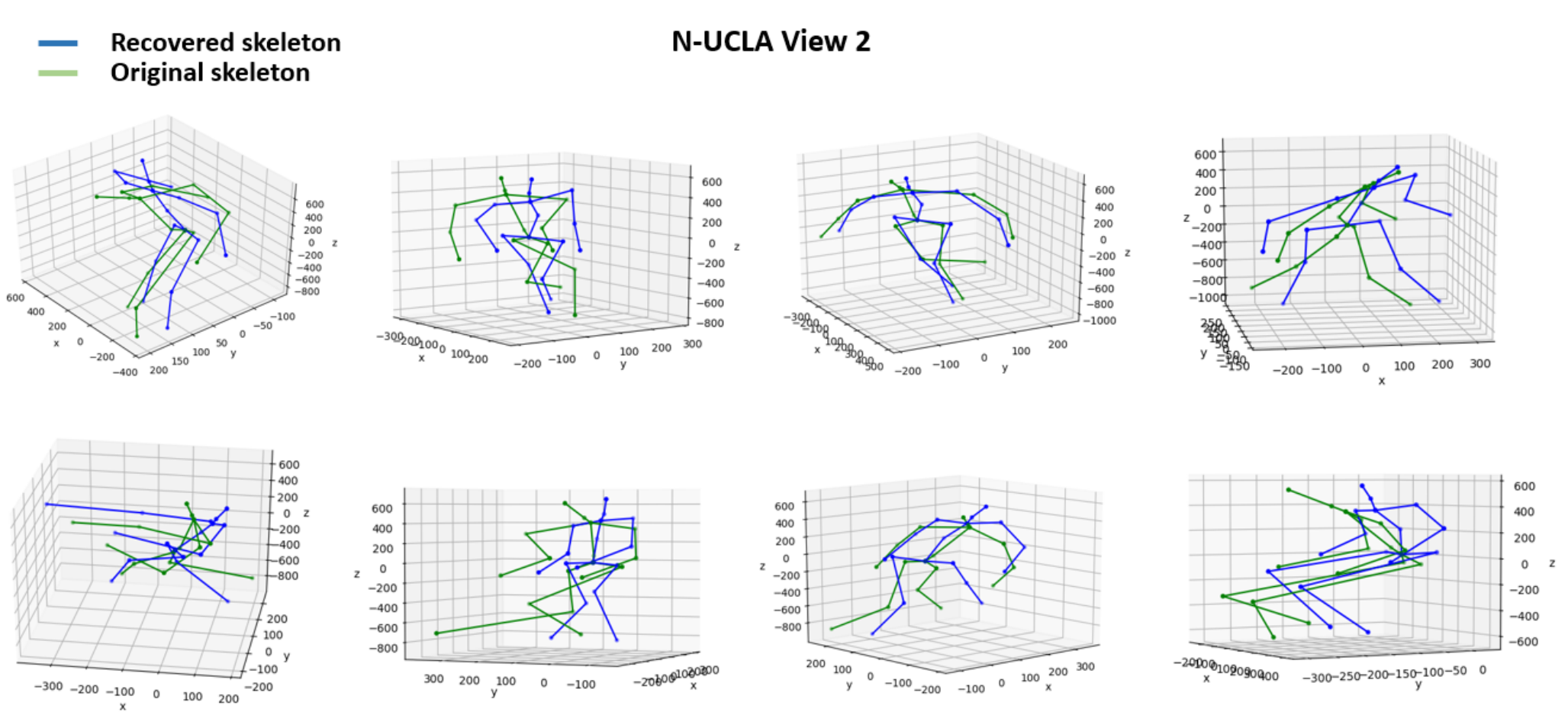}
	\caption{Pose recovery results on N-UCLA dataset which is an unseen dataset to our pretrained DAE (Better to view in color mode with scaling)}
	\label{fig:dnucla}       
\end{figure}

To further demonstrate that the learned representation does encode the intrinsic feature of human 3D pose, we applied the pretrained network on unseen N-UCLA dataset for a qualitative pose denoising evaluation. As Fig.~\ref{fig:dnucla} shows, from perspectives of fixed bone length, symmetry, and motion limit of human joints, the recovered skeletons are much more stable and reasonable compared to the original skeletons captured by the 3D sensor. The capability of denoising unseen skeleton verifies that our network has learned the intrinsic feature of human 3D pose. 

\subsection{Evaluation on Unsupervised Cross-View Action Recognition}
To evaluate the learned pose-dependent feature, we further exploit it for unsupervised cross-view action recognition on the N-UCLA dataset and NTU RGB+D dataset. The results are shown in Table~\ref{tab: AC_NUCLA}. In unsupervised action recognition, it's a general way to keep the pre-trained encoder fixed and only train the classifier~\cite{demisse2018pose,li2018unsupervised,luo2017unsupervised}. As our target is to evaluate the performance of learned pose-dependent representation in cross-view action recognition, a simple 1-layer LSTM is adopted as classifier to reduce the influence of classifier design. Also, to this end, we only compare with those state-of-the-art methods based on RNN. Though our classifier is much simpler than those compared methods, the accuracy we achieved is competitive and even surpass some of the supervised methods. Action recognitions that are directly based on skeleton coordinates are used as baselines. Among them, the "raw coordinates" means directly feeding the raw coordinates of skeletons into classifier. The "normalized coordinates" means the poses are further normalized according to the mean position and standard deviation of joints. Translation of human pose is neglected when training the DAE. But for action recognition, the translation, which should be a part of the human motion, is concatenated together with the learned pose-dependent feature.

\begin{table}[t!]
	\caption{Results of the cross-view action recognition on the N-UCLA dataset and the NTU RGB+D dataset (* means the result is reproduced by implementing the model reported)}
	\label{tab: AC_NUCLA}
	\begin{center}
		\begin{tabular}{l|l|lcc}
			\hline\noalign{\smallskip}
			Dataset &         & Method & Acc.(\%)   &\# of params.\\
			\hline\noalign{\smallskip}		
			\multirow{9}{*}{N-UCLA}  &\multirow{3}{*}{Baselines} & raw coordinates  &38.72 & -\\
			&         &normalized coordinates  &48.69 & -\\	
			\cline{2-5}
			&\multirow{4}{*}{Supervised} &TLDS~\cite{ding2018tensor} &74.6 & -\\	
			&           & {HBRNN-L~\cite{du2015hierarchical}} & 78.52  & - \\
			&           & {Multi-task RNN~\cite{wang2018learning}} & 87.3 & - \\	
			&           & AGC-LSTM~\cite{si2019attention}      &93.3 & -\\
			\cline{2-5}
			&\multirow{4}{*}{Unsupervised} & Li et al. \cite{li2018unsupervised} & 62.5 & -\\ 
			&             & {LongT GAN~\cite{zheng2018unsupervised}} &74.3* & -\\
			&             & {Denoised-LSTM~\cite{demisse2018pose}} &76.81  & - \\			
			&             & \bf{Ours (1-layer LSTM)}              &\bf{80.30}   & -\\
			\hline\noalign{\smallskip}
			\hline\noalign{\smallskip}
			\multirow{11}{*}{NTU RGB+D} &Baseline &normalized coordinates  &69.08  & -\\	
			\cline{2-5}
			&\multirow{7}{*}{Supervised} &Hand-crafted LARP~\cite{vemulapalli2014human} &52.76  & -\\
			&           &{LieGroups~\cite{huang2017deep}} &66.95  & -\\
			&           &{Part-aware LSTM~\cite{Shahroudy_2016_NTURGBD}} &70.27  & -\\
			&           &{ST-LSTM+TG~\cite{liu2016spatio}} &77.70  & 15.37M\\
			&           &{Two-stream GCA-LSTM~\cite{liu2018skeleton}}	&85.10  & 24.54M\\
			&           &Bayesian GC-LSTM~\cite{zhao2019bayesian}   &89.0  & -\\
			&           &AGC-LSTM~\cite{si2019attention}   &95.0  & $>$10.75M\\
			\cline{2-5}
			&\multirow{3}{*}{Unsupervised}
			& LongT GAN~\cite{zheng2018unsupervised}     &48.1*   & 40.18M\\
			& & EnGAN-PoseRNN~\cite{kundu2019unsupervised}     &77.8   & $>$0.7M\\   
			&  & \bf{Ours (1-layer LSTM)}           &\bf{79.71}   & 0.27M\\			
			\hline\noalign{\smallskip}
		\end{tabular}
	\end{center}	
\end{table}

It shows explicitly in Table.~\ref{tab: AC_NUCLA} that the learned pose-dependent feature improves the cross-view action recognition accuracy significantly compared with baseline results, about improving by 30\% on N-UCLA dataset and 10\% on NTU RGB+D dataset. Among those unsupervised methods on N-UCLA dataset, our method achieves the best performance with an increment of 18\% compared to the work of \cite{li2018unsupervised}. The method of \cite{li2018unsupervised} is exclusively designed for learning a temporal representation using sequential skeletons in action recognition, while our method is designed for learning a representation from single pose. The accuracy of Denoised-LSTM~\cite{demisse2018pose} which is based on conventional DAE is quite close to our result, but the feature they learned is not view-invariant and a preprocessing treatment is needed to alleviate the influence of view changing. A similar performance is reported on NTU RGB+D dataset. Even compared with supervised methods, the accuracy is better than some of them that have more complex classifier. Performance attained on these two benchmark datasets sufficiently demonstrates the effectiveness and robustness of the learned pose representation in our method.

Though temporal information is not considered in learning pose representation, the performance in action recognition indicates that informative temporal features still can be extracted from sequential learned representations with simple LSTM layer, which should be attributed to the intrinsic feature of human pose it has learned.

Moreover, as shown in Table~\ref{tab: AC_NUCLA}, we also contrast the size of model with other state-of-the-art works that evaluated on the NTU dataset. Considering the SeBiReNet and all MLP layers used in our learning architecture, the learnable parameters in our method is about 0.27 million. As some details missed in several works, we can only estimate the lowest number of parameters in those methods, such as EnGAN-PoseRNN~\cite{kundu2019unsupervised} and AGC-LSTM~\cite{si2019attention}. It can be seen that our method achieves a competitive result with the least parameters, which also shows the efficiency of our method from another perspective.

\subsection{Ablation Study}


To evaluate the contribution of each part in the learning architecture, we have an ablation study based on the N-UCLA and NTU RGB+D dataset as shown in Table.~\ref{tab:AATSablation}. In Table~\ref{tab:AATSablation}, the raw DAE means the structure denoted in eq.~\ref{e1}. "FD" means the learned latent feature is disentangled to view-dependent feature and pose-dependent feature as denoted in eq.\ref{e2}. $L_O$ refers to the unit orthogonal matrix constraint on view-dependent feature, as denoted in eq.~\ref{e10}. $L_{f}$ and $L(R*p)$ are the feature loss and reconstruction loss of randomly rotated pose as defined in Sec.~\ref{learning_architecture}. "full architecture" means integrating all the components defined in eq.~\ref{e11} for a better disentanglement and representation learning.

\begin{table}[t!]
	\caption{Ablation study based on the N-UCLA dataset. All components contribute to the overall performance and a better disentanglement}
	\label{tab:AATSablation}
	\vspace{-10pt}
	\begin{center}		
		\begin{tabular}{l >{\centering\arraybackslash}p{2cm} >{\centering\arraybackslash}p{2.5cm}}
			\hline
			\multirow{2}{*}{Model} & \multicolumn{2}{c}{Accuracy(\%)}\\
			\cline{2-3}
			  &   & \\[-1em]
			  &N-UCLA & NTU RGB+D\\
			\hline	
			 baseline (relative coordinates)	&51.53    &69.08\\
			 raw DAE  &58.66    &71.11\\
			 raw DAE + Feature Decomposition (FD)            &60.61    &73.99\\
			 raw DAE + FD + $L_O$  &62.55				&74.46\\
			 raw DAE + FD + $L_O$ + $L_{f}$       &73.81     &75.72\\
			 raw DAE + FD + $L_O$ + $L_{f}$ + $L(R*p)$   &76.84	&77.07\\		
			 full architecture             &\bf{80.30} 			&\bf{79.71} \\
			\hline
		\end{tabular}
	\end{center}	
\end{table}

As shown in Table~\ref{tab:AATSablation}, the raw DAE with skeleton corruption achieved an accuracy of 58.66\%, which is 10\% higher than the baseline result. By disentangling the latent feature and adding orthogonal loss to view-dependent feature, another 4\% improvement is obtained. However, the accuracy steeply increase to 73.81\% when adding the feature loss to pose-dependent feature, which indicates that the network learns better view-invariant pose feature in this case. The reconstruction losses of randomly rotated pose and generated view-transfered poses can further help improve the performance to 80.3\% in cross-view action recognition, which indicates the features are better disentangled. The improvements brought by different components are steady on the NTU RGB+D dataset, but all the components designed in our method contribute to the final performance. Feature loss and view-transferred pose losses are strong regularizations in preserving all the intrinsic pose information and learning view-invariant representations. The results, in turn, demonstrate the effectiveness of disentangling features rather than only extracting the view-invariant feature.

\begin{figure}[t!]
   \begin{tabular}{lr}
   
    \begin{minipage}[m]{0.49\linewidth}
	\centering
	\includegraphics[width=\linewidth]{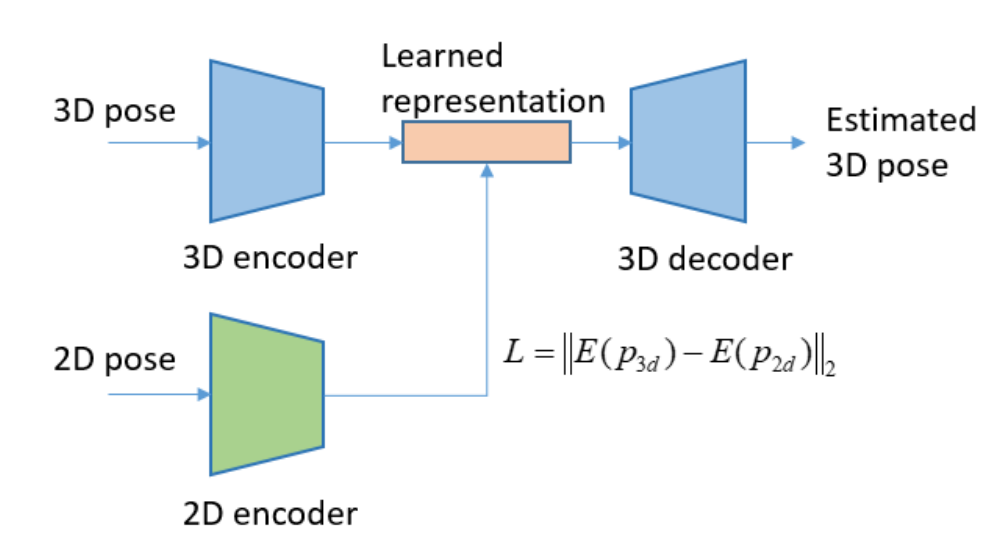}\\
	
	\caption{Extension for 3D pose estimation}
	\label{fig:3dextension}       
	\end{minipage}
	\begin{minipage}[m]{0.48\linewidth}
	\centering
	\makeatletter\def\@captype{table}\makeatother\caption{3D pose estimation from the generated 2D pose of H3.6M dataset}
	\label{tab:3Destimation}
	\begin{tabular}{c|c}
	\hline
	     Method  & MPJPE(mm)  \\
	\hline
	     aGCN~\cite{yang2018graph}    &82.9  \\
	     ST-GCN~\cite{yan2018spatial}  &57.4  \\
	     Martinez et al.~\cite{martinez2017simple}   & 45.5 \\
	     SemGraph~\cite{zhao2019semantic}    & 43.8 \\
	\hline
	    \textbf{Ours}              & 53.1 \\
	\hline
	\end{tabular}
    \end{minipage}
    \end{tabular}
\end{figure}

\subsection{Extension Evaluation on 3D Pose Estimation}
We further design a simple frame to explore the extension of the learned representation for 3D pose estimation from 2D pose. The extension frame contains a 3D encoder, a 2D encoder, and a decoder as shown in Fig.~\ref{fig:3dextension}. The 3D encoder and decoder form a 3D stream and are pre-trained using 3D poses as we did in the former section. Encoder and decoder are the same with DAE in Fig.~\ref{fig:wholearchitecture}. In the second step, by regularizing the 2D encoder to learn a representation similar to the representation obtained in the 3D stream, 3D pose is expected to be estimated from the 2D pose. The result achieved by finetuning the 3D stream on H3.6M dataset as shown in Table~\ref{tab:3Destimation}. It can be seen that the learned representation is also applicable to the 3D pose estimation with a simple frame.

\section{Conclusion}

In this paper, we propose a neural network architecture to learn a human 3D pose representation by disentangling the view-dependent and pose-dependent features. Different from previous methods, the proposed method use the view-dependent and pose-dependent feature together as a pose representation for sake of preserving information. A SeBiReNet is proposed to model the human skeleton data, which considers the kinematic dependency between body joints of the human skeleton. Extensive experiments prove that the learned representation keeps the intrinsic feature of the human 3D pose and is capable of achieving excellent performance in skeleton denoising and unsupervised action recognition tasks. Utilizing the disentangled pose feature, our extension research will be focused on the view transfer between different poses.
\\

\clearpage
%
%
\bibliographystyle{splncs04}
\bibliography{eccv2020_3152}
\end{document}